\documentclass[runningheads]{llncs}
\usepackage[T1]{fontenc}
\usepackage{amsmath}
\usepackage{amssymb}
\usepackage{mathtools}
\usepackage{subcaption}
\usepackage{xspace}
\usepackage{graphicx}
\usepackage{microtype}
\usepackage{graphicx}
\usepackage{booktabs} 

\usepackage{booktabs}
\usepackage{pifont}
\usepackage{algorithm}
\usepackage{algorithmic}
\usepackage{multirow}
\usepackage{url}
\usepackage[capitalize,noabbrev]{cleveref}

\AtBeginDocument{%
  \providecommand\BibTeX{{%
    \normalfont B\kern-0.5em{\scshape i\kern-0.25em b}\kern-0.8em\TeX}}}

\newcommand{\proj}{RaFL\xspace} 

\begin{document}

\title{Resource-Aware Heterogeneous Federated Learning with Specialized Local Models}
\titlerunning{Resource-Aware Heterogeneous Federated Learning}

\author{Sixing Yu$^{1}$, J. Pablo Muñoz$^{2}$, Ali Jannesari$^{1}$}

\authorrunning{Yu et al.}
\institute{$^{1}$Iowa State University,  $^{2}$Intel Labs\\
\email{\{yusx, jannesar\}@iastate.edu, pablo.munoz@intel.com}}


\maketitle

\begin{abstract}
Federated Learning (FL) is extensively used to train AI/ML models in distributed and privacy-preserving settings.
Participant edge devices in FL systems typically contain non-independent and identically distributed~(Non-IID) private data and unevenly distributed computational resources. Preserving user data privacy while optimizing AI/ML models in a heterogeneous federated network requires us to address data and system/resource heterogeneity.
To address these challenges, we propose \underline{R}esource-\underline{a}ware \underline{F}ederated \underline{L}earning~(\proj). \proj allocates resource-aware specialized models to edge devices using Neural Architecture Search~(NAS) and allows heterogeneous model architecture deployment by knowledge extraction and fusion.
Combining NAS and FL enables on-demand customized model deployment for resource-diverse edge devices. Furthermore, we propose a multi-model architecture fusion scheme allowing the aggregation of the distributed learning results.
Results demonstrate \proj's superior resource efficiency compared to SoTA. 

\end{abstract}





\section{Introduction}

Federated Learning (FL) has emerged as a privacy-aware decentralized AI/ML model training and optimizing paradigm, widely adopted by the technology industry, to leverage the massive data generated at the edge (e.g., mobile phones and IoT devices). FL involves local client nodes distributively training and optimizing AI models, with an aggregating server integrating the decentralized results without accessing users' private data.

However, FL introduces two key challenges: data heterogeneity and resource/system heterogeneity, making it difficult to scale. The private data collected on edge devices is non-independent and identically distributed (Non-IID), causing uncertainties and optimization failures when training with naïve decentralized methods~\cite{Kairouz2021problemsfl}. Additionally, resource heterogeneity, where edge devices have different specifications and computational capacities, leads to inefficient resource utilization. Moreover, the expensive communication overhead produced by frequently sharing weights/gradients between edge clients and servers becomes a bottleneck for scaling up the network, especially in cross-device and model-centric FL settings involving possibly millions of edge devices, resulting in massive bandwidth consumption.

State-of-the-art (SoTA) FL baselines mainly focus on addressing data heterogeneity (i.e., non-IID data on edge devices), while ignoring system and resource heterogeneity among edge devices, which exhibit significant variations, particularly in cross-device FL settings. Simply deploying the same model architecture to every client leads to poor resource utilization, causing resource-hungry edge devices to consume precious energy supplies while underutilizing more capable edge devices. Although AutoML solutions, such as Neural Architecture Search (NAS)~\cite{bootstrapNAS,cai20ofa}, have achieved success in generating high-performance neural architecture models for resource-heterogeneous environments, integrating NAS into existing FL solutions presents challenges, as it requires identical neural architecture among clients for aggregating decentralized training results. Recent efforts~\cite{he2020fednas,yu2021adaptive_pruning} dedicated to integrating NAS, network pruning, and weights dropout in FL rely on zero-padding model weights to enable multi-neural architectures, helping in model sparsity but offering limited contributions to model size reduction and communication efficiency.

To address both data and system heterogeneity, we propose Resource-aware Federated Learning (\proj), a multi-architecture FL framework. \proj leverages a weight-sharing supernet to efficiently specialize resource-aware models on edge devices, tailored to their specific resource constraints. This approach enables the deployment of diverse model architectures across edge-FL clients while maintaining the ability to aggregate and share knowledge.
Central to \proj is the use of a small-size \textbf{knowledge network} that cooperates with the resource-aware local model for neural knowledge sharing among clients. \proj clients engage in deep mutual learning~\cite{zhang18dml} to co-train their network pairs and diffuse knowledge into their knowledge networks. 
The \proj server then aggregates the local knowledge from each client's knowledge network, effectively combining the decentralized training results and supporting multi-architecture FL.
Given the availability of public data, \proj provides the option of using ensemble distillation to improve the robustness of knowledge fusion.
\proj makes the following contributions:
\begin{itemize}
\item Mitigates resource/system heterogeneity by deploying resource-aware neural architectures, and maximizes resource utilization.

\item Provides ensemble knowledge distillation and transfer learning algorithms specifically designed for federated learning, which aim to improve the robustness of knowledge fusion.

\item Employs high-performing specialized neural architectures to accelerate inference at the edge.

\item Reduces FL communication overhead by using a smaller interceding knowledge network.
\item Provides alternative configurations to take advantage of transfer learning, and public data, making it scalable and applicable in real-world scenarios.

\end{itemize}

\section{Related Works}

\label{sec:relat}

\noindent\textbf{Federated Learning}
The pioneering FL research by~\cite{mcmahan2017fedavg} introduced FedAvg, which combined local stochastic gradient descent (SGD) model optimization with global model averaging. 
Recently, several variants of FedAvg have emerged. For example, FedProx \cite{li2020fedprox} was proposed to address convergence issues associated with data and device heterogeneity, which was achieved by permitting the participation of ``straggler'' devices and introducing a proximal term in the local loss. FedNova \cite{wang2020fednova}  normalized and scaled local updates by modifying weights to mitigate gradient bias. SCAFFOLD \cite{karimireddy2020scaffold} introduced gradient-based control variates to correct client drift and speed up convergence. 
These SoTA methods inherently assume uniformity in edge resource capacities by deploying architecturally identical models onto resource-heterogeneous edge devices.
In contrast, heterogeneous edge computational capacities require the deployment of specialized client network architectures. As such, accounting for potential network architectural diversity is imperative.  

\noindent\textbf{Personalized FL}
Personalized FL~\cite{fallah2020personalized_meta,dinh2021personalized_moreau,hanzely2020lowerbound_personazlied,huang2021personalized_crosssilo,zhang2021personalized_modelopt} has gained attention as a solution to the challenges posed by \textit{statistical heterogeneity}, such as non-IID data distribution, varied sample sizes among clients, and imbalanced class distributions across local datasets. Uniform FL models can degrade overall performance, while personalized models increase robustness.
Personalized FL allows edge clients to tune model weights to better fit local data, but these approaches still adopt the same model architecture at the client, failing to address the heterogeneity of computational capacities.

\noindent\textbf{Knowledge Distillation in Federated Learning}
Knowledge distillation (KD) has been adopted in FL as a means to address model heterogeneity \cite{li19fedmd,he20fedgkt,seo20fedkd,sui20feded,lin2020feddf}. 
For example, Fed-ensemble \cite{sui20feded} integrates the prediction output of all client models; FedKD~\cite{wu22fedkd} proposes an adaptive mutual distillation framework to learn a student and a teacher model simultaneously on the client side; FedDF \cite{lin2020feddf} distills the ensemble of client and teacher models to a server student model; \cite{li19fedmd}, proposes the application of a public dataset as a medium of exchanging knowledge among customized client models;
\cite{yu20salvaging} proposes the use of KD to transfer knowledge between a global network and a student network rather than using a direct assignment.
In contrast, our approach utilizes local deep mutual learning \cite{zhang18dml} coupled with an optional ensemble-based multi-model fusion in the cloud whenever public data is available. In the absence of public data, we perform global model aggregation. This makes our pipeline quite flexible and applicable in real-world scenarios.



\section{Methodology}
\label{sec:method}
\begin{figure*}%
    \vspace{-.5 cm}

    \centering
    \includegraphics[width=.8\linewidth]{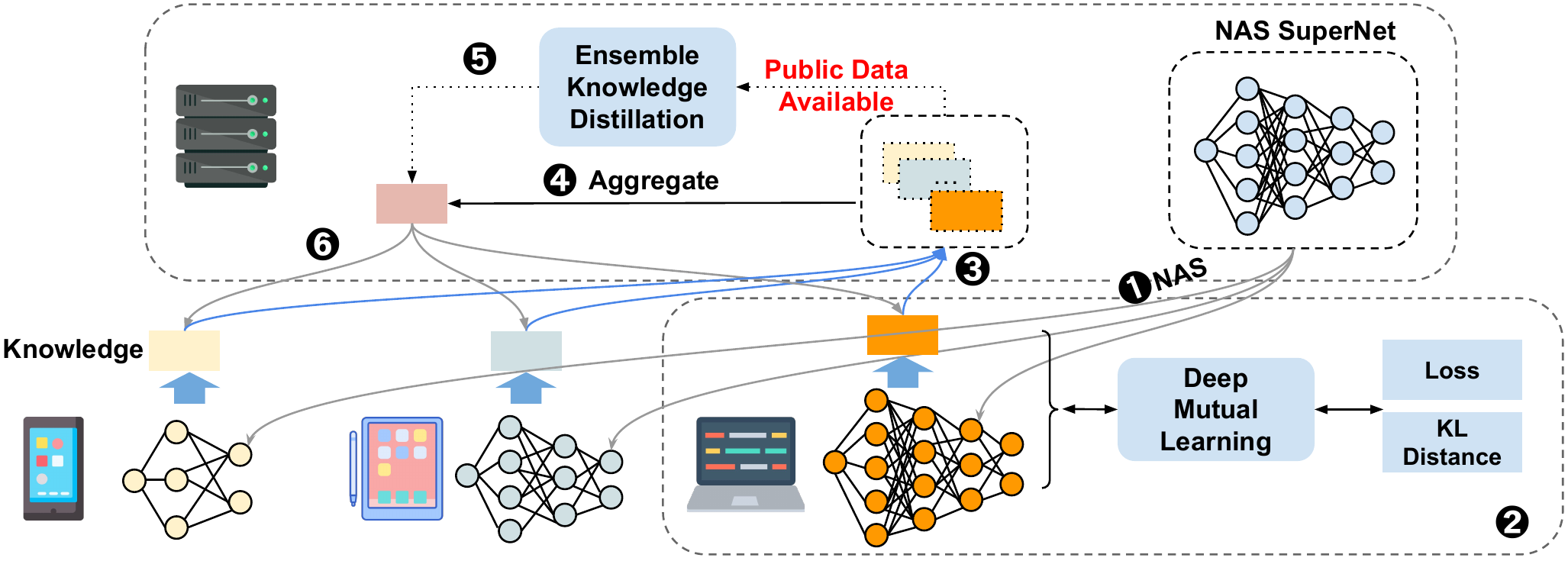}
    \vspace{-.4 cm}

    \caption{During initialization, edge clients request for specialized resource-aware models from the weight-sharing supernet with neural architecture search (NAS). During local training, the local resource-aware model and knowledge network are co-trained together via deep mutual learning. The server fuses the neural knowledge from clients and performs optional ensemble learning if public data is present.  }%
    \label{fig:overview}%
        \vspace{-.5cm}

\end{figure*}
There are three main stages in \proj:  resource-aware model specialization, local knowledge fusion with deep mutual learning, and cloud knowledge aggregation (Figure~\ref{fig:overview}).
\proj performs an on-demand model architecture search from weight-sharing supernet~\cite{cai20ofa} to tailor to the resource utilization requirements of edge devices, unlike mainstream FL systems that deploy identical model architectures on all devices (\ding{202} in Figure~\ref{fig:overview}). Local devices train their specialized model on local data and transfer the model's knowledge to a smaller knowledge network via deep mutual learning (\ding{203}). Edge clients then communicate their local knowledge to the cloud server (\ding{204}), which aggregates the received knowledge into a global knowledge network (\ding{205}). When public data is available, \proj optionally provides ensemble distillation to further improve the robustness of the knowledge aggregation stage (\ding{206}). Finally, the \proj server transfers the global knowledge to the edge devices (\ding{207}).

\subsection{Resource-aware Federated Learning using Specialized Models}

Mainstream FL solutions assume homogeneous resource capacities across clients, which is impractical in cross-device settings with varying computational resources. Simply deploying an identical model architecture leads to poor resource utilization. \proj proposes a resource-aware federated neural architecture search to obtain resource-tailored models for heterogeneous edge devices.

We first deploy a weight-sharing super-network on the cloud. During FL initialization, clients query this supernet~\cite{cai20ofa} to obtain resource-aware subnets matched to their computational constraints. The supernet is trained by minimizing a validation loss over a distribution of subnet architectures (Equation \ref{eq:nasobj}).

\begin{equation}
\label{eq:nasobj}
\min_{\Theta}\sum_{arch_i} L_{val}(C(\Theta, arch_i))
\end{equation}

Where $\Theta$ represents the super-network weights, $arch_i$ is a subnetwork configuration, and $C(\Theta, arch_i)$ samples the subnetwork weights from the super-network. Once trained, subnetworks can be efficiently derived without retraining the super-network.

In the NAS process, \proj defines a search space $\mathcal{S}$ that encompasses the possible architectures for the specialized models. The search space is constrained by a set of resource requirements $\mathcal{R}$, which includes memory, computational power, and energy consumption limitations of the edge devices. The objective is to find an optimal architecture $arch^*$ that maximizes performance while satisfying the resource constraints:

\begin{equation}
arch^* = \arg\max_{arch \in \mathcal{S}} P(arch) \quad \text{s.t.} \quad R(arch) \leq \mathcal{R}
\end{equation}

where $P(arch)$ represents the performance metric of the architecture $arch$, and $R(arch)$ denotes the resource consumption of $arch$. \proj employs a search strategy to efficiently explore the search space and extract the specialized models $M_i$ from the supernet, tailored to the specific capabilities of each edge device $i$. This process ensures that the obtained models maximize resource utilization and performance in the federated learning setting.

\subsection{Local Knowledge Fusion}

Traditional FL methods deploy a uniform model across clients, enabling simple weight/gradient averaging for aggregation. However, in \proj, where varying resource-aware architectures are used, weight averaging is infeasible. Instead, we propose communicating extracted knowledge from local models via deep mutual learning (DML) \cite{zhang18dml}.

For the $l^{th}$ client $C_l$, we denote its resource-aware model as $\theta_l$ and a downloaded knowledge network as $\theta_l^k$. $C_l$ jointly optimizes $\theta_l$ and $\theta_l^k$ on its local data. For an input batch, we first computes the cross-entropy loss $L_c(\theta;B)$ (Equation \ref{eq:ce}) for both networks independently.

\begin{equation}
\label{eq:ce}
L_c(\theta;B) = -\frac{1}{|B|}\sum_{x,y \in B} y^T \log(\sigma(\theta(x)))
\end{equation}

It then measures the Kullback-Leibler (KL) divergence $D_{KL}(\theta_l||\theta_l^k;B)$ (Equation \ref{eq:kl1}) between the predicted distributions.

\begin{equation}
\label{eq:kl1}
D_{KL}(\theta_{l}||\theta_{l}^k; B) = \frac{1}{|B|}\sum_{x,y\in B} \sigma(\theta_{l}(x)^T) \log(\frac{\sigma(\theta_{l}(x))}{\sigma(\theta_{l}^k)})
\end{equation}

The mutual learning loss (Equations \ref{eq:loss1} and \ref{eq:loss2}) combines the cross-entropy loss and KL divergence, enabling learning from data while aligning the networks.

\begin{equation}
\label{eq:loss1}
L_{\theta_{l}} = L_c(\theta_l;B) + D_{KL}(\theta_{l}^k||\theta_{l};B)
\end{equation}

\begin{equation}
\label{eq:loss2}
L_{\theta_{l}^k} = L_c(\theta_l^k;B) + D_{KL}(\theta_{l}||\theta_{l}^k;B)
\end{equation}

DML outperforms solo learning due to the dissimilarity between the participating models ($\theta_l$ and $\theta_l^k$). Their varying architectures and the intermittent aggregation of $\theta_l^k$ enable learning different data representations, allowing them to capture each other's knowledge. Moreover, while the smaller $\theta_l^k$ reduces communication overhead, co-learning with the larger $\theta_l$ boosts its performance.

As clients communicate their $\theta_l^k$ networks, they share local knowledge and receive global knowledge. Successive DML steps further converge both $\theta_l$ and $\theta_l^k$ on the global data distribution. Overall, initializing clients with resource-specific NAS subnetworks paired with a smaller $\theta_l^k$ addresses device heterogeneity and reduces communication overhead.
\subsection{Cloud Knowledge Aggregation and Ensemble Distillation}
\label{sec:Cloud}

\proj utilizes the knowledge network as a medium to support interoperability among multi-model FL clients. The function of the knowledge network is to exchange knowledge among Non-IID clients (Figure~\ref{fig:overview} \ding{204} and \ding{207}). \proj provides two model fusion solutions, weight aggregating and ensemble knowledge distillation, to support comprehensive FL practical scenarios (Figure~\ref{fig:overview} \ding{205} and \ding{206}). 

Assume $\theta^k$ is the global knowledge network, and $\theta^k_l$ as the $l^{th}$ client's knowledge network. We select a set of communication clients $S$ in the current communication round. Once we receive the knowledge from edges, we aggregate the knowledge networks (Equation \ref{eq:aggre}).

\begin{equation}
\label{eq:aggre}
\theta^k \leftarrow \frac{n_l}{N_S} \sum_{l \in S} \theta_l^k
\end{equation}

Here, $N_S$ is the total data size in selected clients $S$, and $n_l$ is the number of data samples in the $l^{th}$ client.
When public data is available, \proj provides an ensemble knowledge distillation option for boosting cloud aggregation. \proj ensembles knowledge networks using the average logits strategy (Equation~\ref{eq:ens}) to produce a combined output on the public dataset. The outputted soft labels are then coupled with the unlabeled dataset to train the global knowledge network in tandem. The distillation loss is defined in Equation~\ref{eq:dist}.

\begin{equation}
\label{eq:ens}
\theta_{ens}(x) = Avg(\theta_{l}^k(x))_{l\in S}
\end{equation}

\begin{equation}
\label{eq:dist}
L_d = D_{KL}(\theta_{ens}|| \theta^k; B)
\end{equation}

Notably, the knowledge network is a small-sized network derived from the NAS super-network. Our ablation study shows the trade-off between the size and the overall performance of the knowledge network, indicating no significant gain in increasing the size of the knowledge network.

\section{Experiments}
\label{sec:exp}

\subsection{Experimental Setup}
\label{sec:setup}
\noindent\textbf{Datasets and models.} We conducted experiments on datasets consistent with the baselines: CIFAR-10/100~\cite{krizhevsky2009cifar}, FEMNIST~\cite{caldas2019leaf}
under Non-IID benchmark settings~\cite{li2022niidbench}. 
The models we deployedwere sampled from MobileNetV2/V3~\cite{howard18mobilenetv2,howard19mobilenetv3} or ResNet \cite{he2016resnet} super-networks \cite{cai20ofa}. 
To avoid confusion, we identify networks by their FLOPs, e.g., we identify ResNet-34 as ResNet with $76$ MFLOPs.
\\
\textbf{Federated Learning settings.} 
We set up different numbers of clients from $30$ to $3000$ in our experimental FL environment, with $10\%$ to $70\%$ client participation rate in each round of communication. We simulated scaled, dynamic sporadic, and asynchronous FL scenarios. Clients are allocated to Non-IID local datasets following the benchmark's Non-IID settings~\cite{li2022niidbench}. 
\\
\textbf{NAS settings.} We initialize an weight-sharing supernet with architectures for ResNet, MobileNetV2, and MobileNetV3. 
\\
\textbf{Baselines.}
We compare \proj with state-of-the-art (SoTA) FL algorithms, including:
\begin{itemize}
    \item Strong FL optimization methods, such as FedAvg \cite{mcmahan2017fedavg}, FedProx \cite{li2020fedprox}, FedNova \cite{wang2020fednova}, SCAFFOLD~\cite{karimireddy2020scaffold}, SPATL~\cite{yu22spatl}. 
    \item Knowledge distillation-based methods, such as FedDF~\cite{lin2020feddf}.
    \item Neural architecture search in FL methods, such as FedNAS~\cite{he2020fednas}.
\end{itemize}



\begin{figure*}[th]
 \begin{center}
\vspace{-.7 cm}

\centerline{\includegraphics[width=\linewidth]
{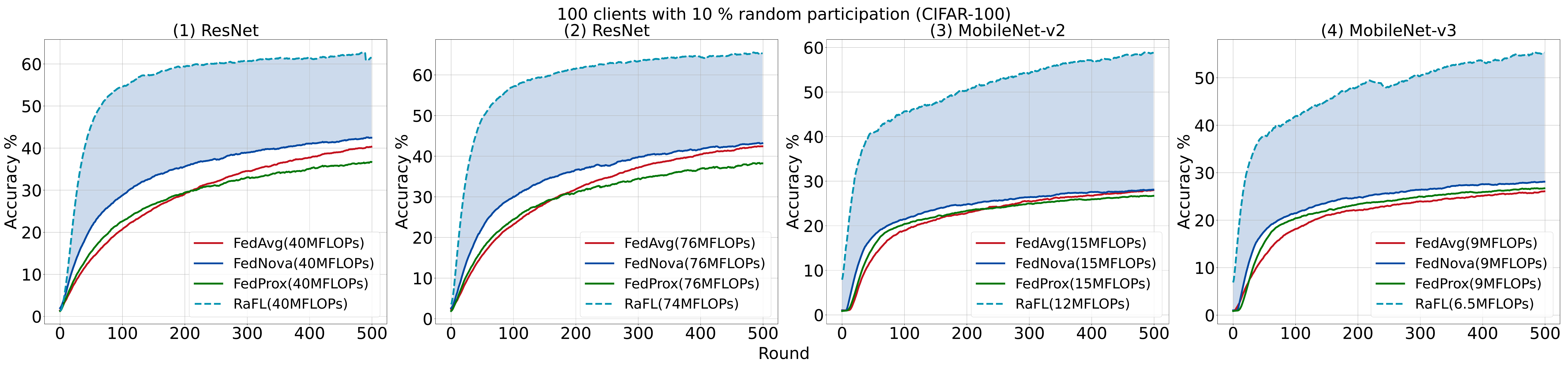}
}
\vspace{-.3 cm}

\caption{Comparison of \proj with SoTAs under \textbf{vanilla aggregation} setting.}
\vspace{-1.4 cm}

\label{fig:vanilla}
\end{center}
 \end{figure*}

\subsection{Learning Efficiency}
\label{sec:learning_efficiency}
We explore the learning efficiency and optimization ability of \proj by evaluating the training performance with respect to communication rounds and comparing it with FedAvg~\cite{mcmahan2017fedavg}, FedNova~\cite{wang2020fednova}, and FedProx~\cite{li2020fedprox}. We consider all possible practical situations ((a), (b), and (c) settings in below) for a fair and extensive comparison.

\begin{figure}[th]
\begin{center}
\begin{subfigure}{.35\textwidth}
\centering
\includegraphics[width=\linewidth, height=5cm, keepaspectratio]{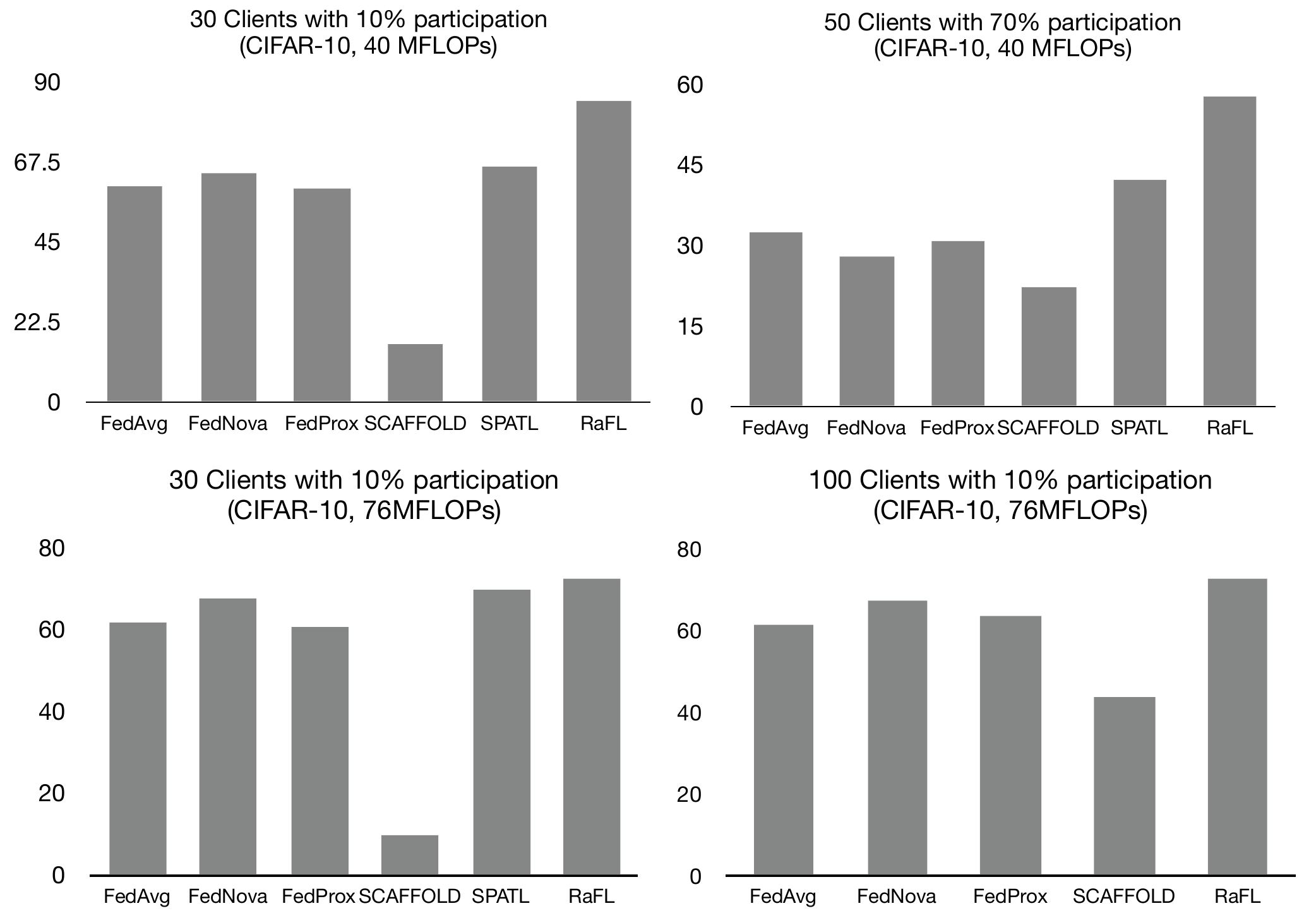}
\caption{Convergence accuracy overhead under various FL settings.}
\label{fig:converge_acc}
\end{subfigure}
\hfill
\begin{subfigure}{.55\textwidth}
\centering
\includegraphics[width=\linewidth, height=5cm, keepaspectratio]{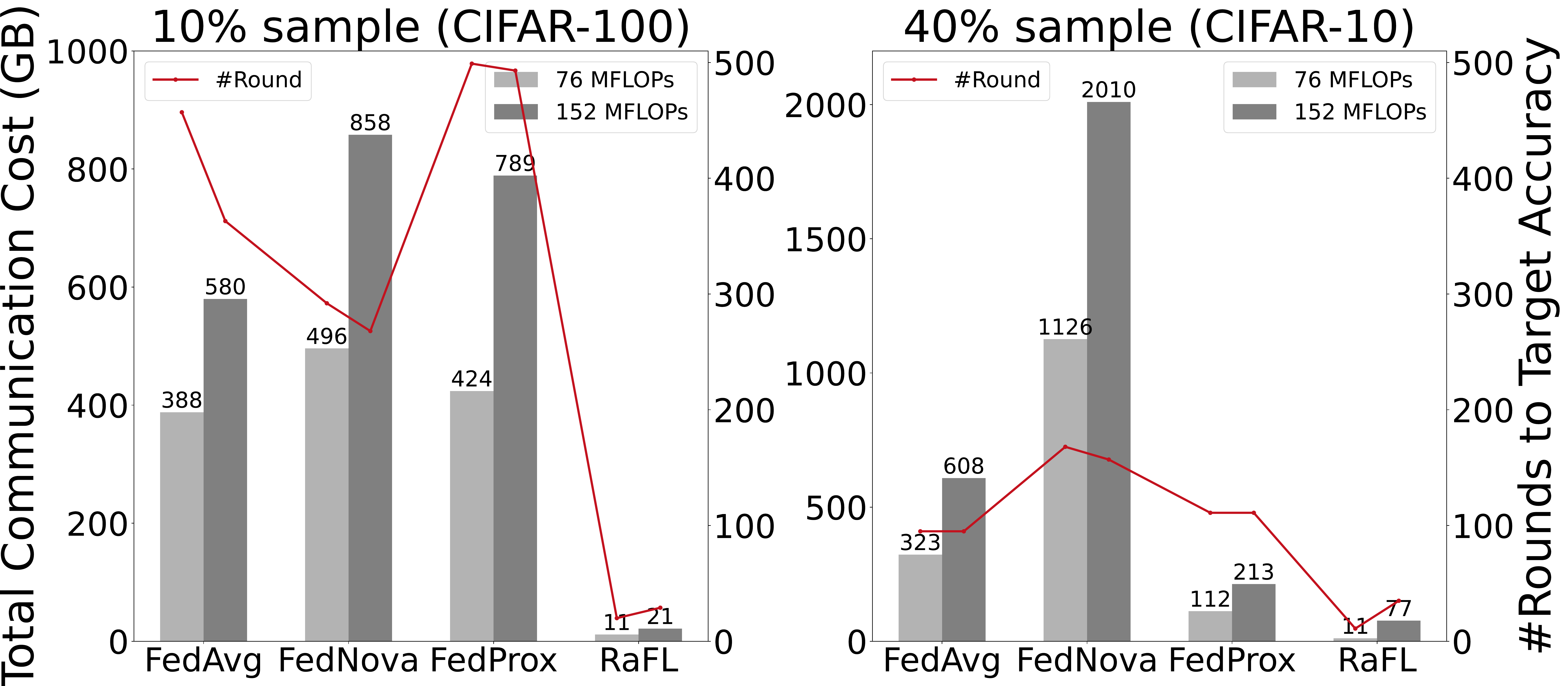}
\caption{Communication cost to achieve target accuracy.}
\label{fig:comm_target}
\end{subfigure}
\caption{Performance comparison of \proj and baselines.}
\label{fig:performance_comparison}
\vspace{-1.5cm}
\end{center}
\end{figure}

\begin{figure}[h]
\begin{center}
\begin{subfigure}{.48\textwidth}
\centering
\includegraphics[width=\linewidth]{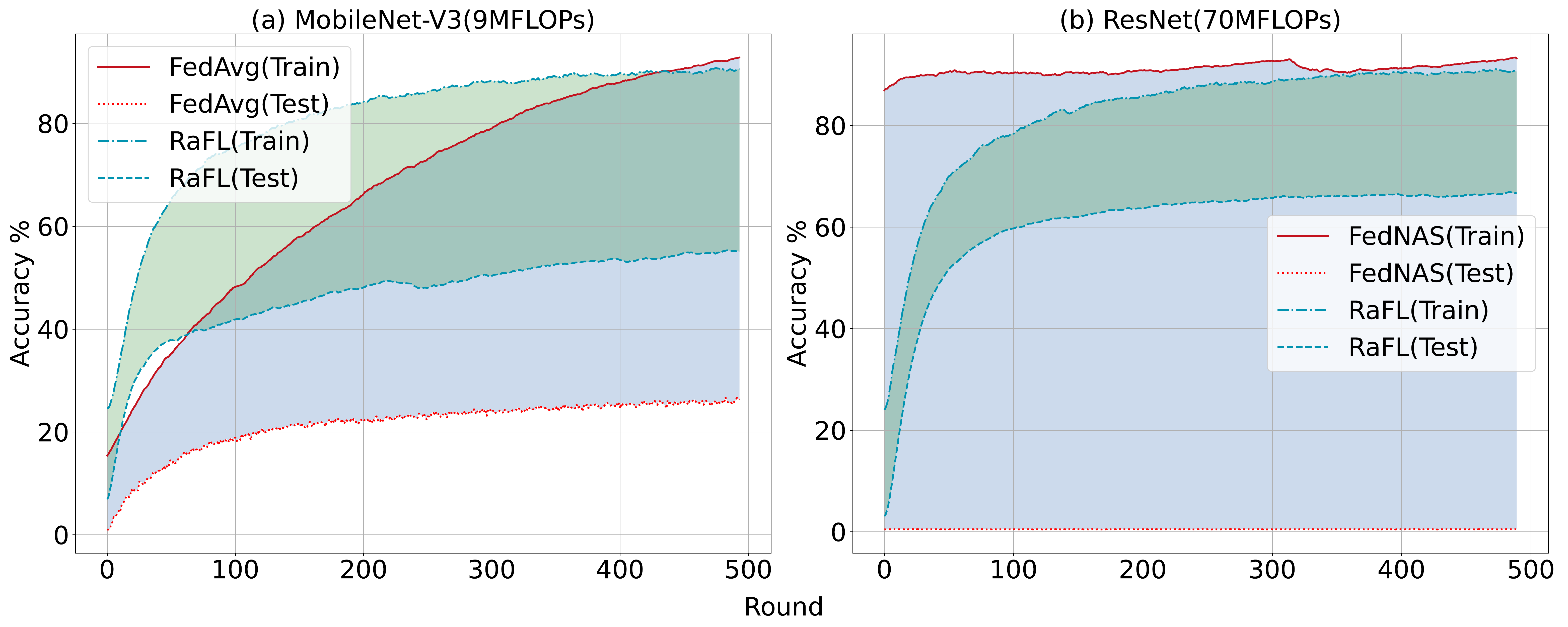}
\caption{Local training accuracy vs. test accuracy.}
\label{fig:overfitting}
\end{subfigure}
\hfill
\begin{subfigure}{.4\textwidth}
\centering
\includegraphics[width=\linewidth]{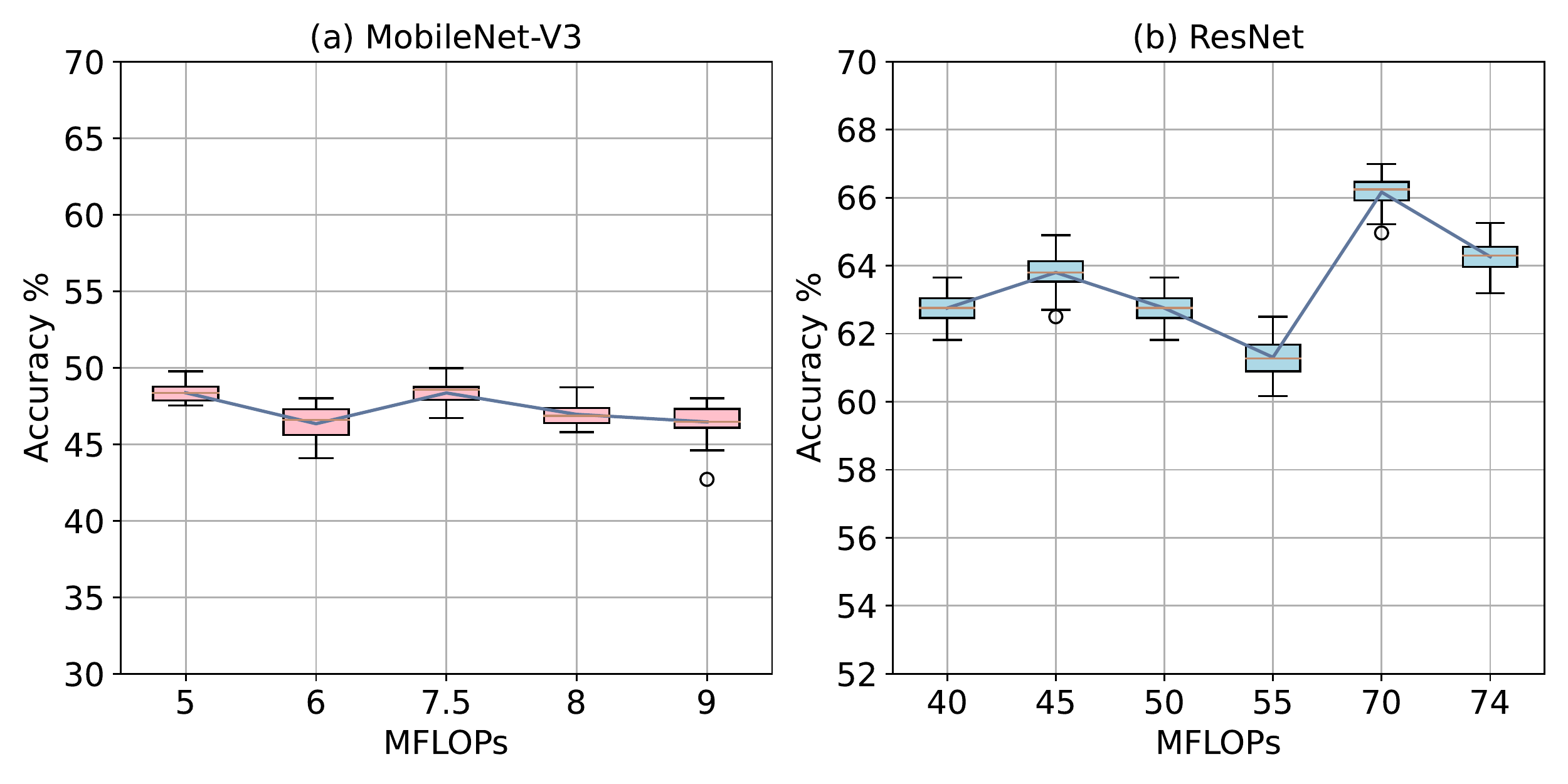}
\caption{\proj under resource heterogeneous setting.}
\label{fig:resource_heto}
\end{subfigure}
\vspace{-.2cm}
\caption{Performance analysis of \proj.}
\label{fig:performance_analysis}
\end{center}
\vspace{-1.2cm}
\end{figure}

\noindent\textbf{(a) Vanilla aggregation.} \proj with vanilla aggregation (Figure~\ref{fig:overview} step \ding{205} and Equation~\ref{eq:aggre}) is used to aggregate local models in each communication round without requiring public data. Figure~\ref{fig:vanilla} shows the comparison results on training performance versus communication rounds under different model architectures and capacities. \proj outperforms the baselines with a large margin and exhibits a stable optimization process while consuming fewer resources. For example, when optimizing MobileNet-v3, \proj produces 28\% higher accuracy than the baselines. Figure~\ref{fig:converge_acc} illustrates that \proj consistently outperforms the baselines across diverse settings and achieves higher convergence accuracy. \proj effectively resists overfitting and shows better learning capacity (Figure~\ref{fig:overfitting}).

\begin{figure*}[th]
 \begin{center}
\vspace{-.2 cm}

\centerline{\includegraphics[width=\linewidth]
{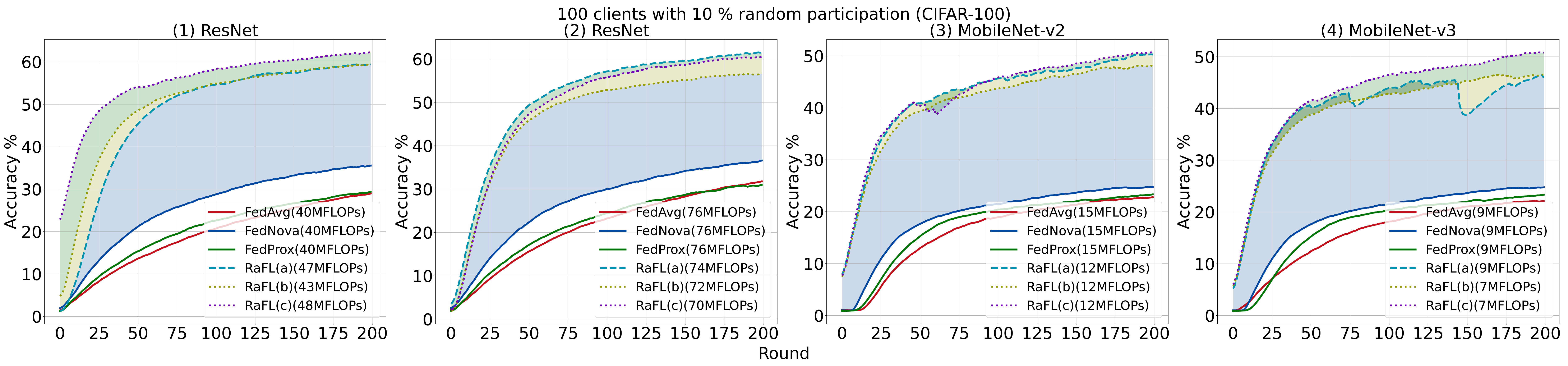}
}
\vspace{-.2 cm}
\caption{Comparison of SoTAs with \proj using \textbf{(a) vanilla aggregation}, \textbf{(b) ensemble distillation} and \textbf{(c) transfer learning} settings.}
\vspace{-.9 cm}
\label{fig:variant}
\end{center}
 \end{figure*}
\noindent\textbf{(b) Public data with ensemble and knowledge distillation.} \proj offers an ensemble distillation option to enhance server aggregation (steps \ding{205} and \ding{206}). Figure~\ref{fig:variant} RaFL(b) shows that \proj equipped with ensemble distillation outperforms strong baselines with significant margins. Ensemble distillation improves and stabilizes training in sporadic connected FL (Figure~\ref{fig:variant} (4)). However, its effectiveness depends on the similarity of public and private client data distributions.

\noindent\textbf{(c) Transfer learning with NAS.} \proj takes advantage of transfer learning on its pre-trained NAS subnetworks when specialized networks are applied to the Non-IID FL dataset. Figure~\ref{fig:variant} RaFL(c) shows that \proj(c) benefits from pre-trained initial weights, resulting in high accuracy in the initial training stage and quick convergence during fine-tuning.


\begin{table*}
\vspace{-.5cm}
\centering
\caption{Communication cost to achieve target accuracy.}
\label{tab:target_acc}
\resizebox{\textwidth}{!}{%
\begin{tabular}{ccclccccll} 
\hline
\multirow{2}{*}{\textbf{Method} }         & \multirow{2}{*}{\textbf{Model Arc.} } & \multirow{2}{*}{\begin{tabular}[c]{@{}c@{}}\textbf{Resource Usage}\\\textbf{ (MFLOPs)} \end{tabular}} & \multirow{2}{*}{\textbf{Clients}} & \multirow{2}{*}{\textbf{Dataset} } & \multirow{2}{*}{\textbf{Target Accuracy} } & \multicolumn{4}{c}{\textbf{Communication Cost} }                                                                                             \\
                                          &                                       &                                                                                                          &                                   &                                    &                                            & \textbf{Round/Client}  & \textbf{Total}              & \multicolumn{1}{c}{\textbf{$\Delta$ Cost} } & \multicolumn{1}{c}{\textbf{Speed Up} }  \\ 
\hline
\multirow{4}{*}{FedAvg~\cite{mcmahan2017fedavg}}  & \multirow{4}{*}{ResNet}               & 40                                                                                                       & \multirow{4}{*}{100}              & \multirow{2}{*}{CIFAR100}          & \multirow{2}{*}{40\%}                      & 0.85 GB                & 388 GB                      & 0GB                                         & (1 $\times$)           \\
                                          &                                       & 76                                                                                                       &                                   &                                    &                                            & 1.6 GB                 & \multicolumn{1}{l}{580 GB}  & 0GB                                         & (1 $\times$)           \\
                                          &                                       & 40                                                                                                       &                                   & \multirow{2}{*}{CIFAR10}           & \multirow{2}{*}{60\%}                      & 3.4 GB                 & 323 GB                      & 0GB                                         & (1 $\times$)           \\
                                          &                                       & 76                                                                                                       &                                   &                                    &                                            & 6.4 GB                 & \multicolumn{1}{l}{608 GB}  & 0GB                                         & (1 $\times$)           \\ 
\hline
\multirow{4}{*}{FedNova~\cite{wang2020fednova}} & \multirow{4}{*}{ResNet}               & 40                                                                                                       & \multirow{4}{*}{100}              & \multirow{2}{*}{CIFAR100}          & \multirow{2}{*}{40\%}                      & 1.7 GB                 & 496 GB                      &                          +108 GB                   &                             (0.78 $\times$)            \\
                                          &                                       & 76                                                                                                       &                                   &                                    &                                            & 3.2 GB                 & \multicolumn{1}{l}{858 GB}  &                 +278 GB                            &                (0.68 $\times$)                         \\
                                          &                                       & 40                                                                                                       &                                   & \multirow{2}{*}{CIFAR 10}          & \multirow{2}{*}{60\%}                      & 6.7 GB                 & 1126 GB                     &                     +803 GB                        &                 (0.29 $\times$)                        \\
                                          &                                       & 76                                                                                                       &                                   &                                    &                                            & 12.8 GB                & \multicolumn{1}{l}{2010 GB} &                          +1402 GB                   &                       (0.30 $\times$)                  \\ 
\hline
\multirow{4}{*}{FedProx~\cite{li2020fedprox}} & \multirow{4}{*}{ResNet}               & 40                                                                                                       & \multirow{4}{*}{100}              & \multirow{2}{*}{CIFAR100}          & 38\%                                       & 0.85 GB                & 424 GB                      &                             +36 GB                &                        (0.92 $\times$)                 \\
                                          &                                       & 76                                                                                                       &                                   &                                    & 40\%                                       & 1.6 GB                 & \multicolumn{1}{l}{789 GB}  &                          +209 GB                   &                          (0.74 $\times$)               \\
                                          &                                       & 40                                                                                                       &                                   & \multirow{2}{*}{CIFAR 10}          & \multirow{2}{*}{60\%}                      & 3.4 GB                 & 377 GB                      &                                 +54 GB            &                    (0.86 $\times$)                     \\
                                          &                                       & 76                                                                                                       &                                   &                                    &                                            & 6.4 GB                 & \multicolumn{1}{l}{710 GB}  &                       +102 GB                      &                 (0.86 $\times$)                        \\ 
\hline
\multirow{3}{*}{RaFL (Ours)}              & \multirow{3}{*}{ResNet}               & 28                                                                                                       &           \multirow{3}{*}{100}                          & \multirow{2}{*}{CIFAR100}          & \multirow{2}{*}{40\%}                      & 0.66 GB                & 13.2 GB                     &                          -374.8 GB                   &                         \textbf{(29.4 $\times$)}                \\
                                          &                                       & 51                                                                                                       &            &                                    &                                            & 1.07 GB                & \multicolumn{1}{l}{31.0 GB} &                                     -549 GB        &                \textbf{(18.70 $\times$)}                         \\
                                          &                                       & 28                                                                                                       &                                   & CIFAR10                            & 60\%                                       & 2.2 GB                 & 77 GB                       &                 -246 GB                            &           \textbf{(4.19 $\times$)}                             \\
\hline
\end{tabular}
}
\vspace{-.8cm}

\end{table*}

\subsection{Communication Efficiency}

\proj exhibits significant superiority in communication efficiency compared to SoTAs, attributed to utilizing smaller knowledge networks for communication. We demonstrate \proj's communication efficiency by optimizing models to reach target performance and achieve converged performance, considering the vanilla aggregation setting (Section \ref{sec:learning_efficiency}.a).

\noindent\textbf{Optimizing Model to Achieve Target Performance.}Table~\ref{tab:target_acc} shows that \proj consistently outperforms SoTAs in communication overhead under various configurations. For example, when using ResNet on CIFAR-100, \proj reduces communication cost by up to $29.4 \times$ compared to FedAvg~\cite{mcmahan2017fedavg} while using significantly less computational resources. Figure~\ref{fig:comm_target} further demonstrates \proj's superiority on larger model capacities (76 MFLOPs and 152 MFLOPs), requiring fewer communication rounds to achieve target performance.

\proj's communication efficiency stems from two main factors: (1) high-performance neural architectures generated using Neural Architecture Search, enabling faster convergence, and (2) knowledge network communication, resulting in a reduced communication cost with up to $4\times$ less bandwidth per round (Table~\ref{tab:target_acc}).

\noindent\textbf{Optimizing Model for Convergent Performance.} Table~\ref{tab:converged_acc} summarizes the results on CIFAR-100, showing that \proj uses less communication cost while achieving significantly better model performance. For instance, \proj reaches around $65 \%$ accuracy on ResNet-74MFLOPs, while the next best method, FedNova, converges at around $43 \%$ with a total cost $1200$ GB higher.

Overall, \proj results in a lower communication cost to achieve both target performance and overall convergence across a wide range of FL settings. Additionally, the model deployed by \proj requires fewer FLOPs at the edge and demonstrates lower resource consumption while achieving higher accuracy, enabling faster inference and better downstream task applications.

\begin{table*}
\vspace{-.5cm}
\centering
\caption{Communication cost to model converge}
\label{tab:converged_acc}
\resizebox{\textwidth}{!}{
\begin{tabular}{cclclcccc} 
\hline
\textbf{Method}                           & \textbf{Model}          & \multicolumn{1}{c}{\textbf{datasets}} & \begin{tabular}[c]{@{}c@{}}\textbf{\textbf{Resource Usage}}\\\textbf{\textbf{ (MFLOPs)}} \end{tabular} & \textbf{Clients}     & \textbf{Round Cost~} & \textbf{Total Cost}  & \textbf{Converged Accuracy}  & \textbf{$\Delta$ Accuracy}  \\ 
\hline
\multirow{4}{*}{FedAvg~\cite{mcmahan2017fedavg}}  & \multirow{2}{*}{ResNet} & \multirow{4}{*}{CIFAR-100}            & 40 MFLOPs                                                                                              & \multirow{4}{*}{100} & 0.85 GB              & 425GB                & (40.00$\pm$1.21)\%           & 0\%                \\
                                          &                         &                                       & 76 MFLOPs                                                                                              &                      & 1.6 GB               & 800GB                & (42.21$\pm$ 0.80)\%          & 0\%                \\
                                          & MobileNet-V2            &                                       & 15 MFLOPs                                                                                              &                      & 280MB                & 137GB                & (27.80$\pm$0.80)\%           & 0\%                \\
                                          & MobileNet-V3            &                                       & 9 MFLOPs                                                                                               &                      & 230MB                & 112GB                & (25.89$\pm$0.87)\%           & 0\%                \\ 
\hline
\multirow{4}{*}{FedNova~\cite{wang2020fednova}} & \multirow{2}{*}{ResNet} & \multirow{4}{*}{CIFAR-100}            & 40 MFLOPs                                                                                              & \multirow{4}{*}{100} & 1.7 GB               & 850GB                & (42.29$\pm$1.23)\%           & +2.29\%            \\
                                          &                         &                                       & 76 MFLOPs                                                                                              &                      & 3.2 GB               & 1600GB               & (43.02$\pm$0.76)\%           & +0.81\%            \\
                                          & MobileNet-V2            &                                       & 15 MFLOPs                                                                                              &                      & 560MB                & 273GB                & (27.98$\pm$0.55)\%           & +0.18\%            \\
                                          & MobileNet-V3            &                                       & 9 MFLOPs                                                                                               &                      & 460MB                & 225GB                & (26.25$\pm$0.82)\%           & +0.36\%            \\ 
\hline
\multirow{3}{*}{FedProx~\cite{li2020fedprox}} & \multirow{2}{*}{ResNet} & \multirow{4}{*}{CIFAR-100}            & 40 MFLOPs                                                                                              & \multirow{4}{*}{100} & 0.85 GB              & 425GB                & (36.43$\pm$1.10)\%           & -3.57\%            \\
                                          &                         &                                       & 76 MFLOPs                                                                                              &                      & 1.6 GB               & 800GB                & (38.04$\pm$1.59)\%           & -4.17\%            \\
                                          & MobileNet-V2            &                                       & 15 MFLOPs                                                                                              &                      & 280MB                & 137GB                & (26.65$\pm$0.82)\%           & -1.15\%            \\
                                          & MobileNet-V3            &                                       & 9 MFLOPs                                                                                               &                      & 230MB                & 112GB                & (24.39$\pm$0.52)\%           & -1.50\%            \\ 
\hline
\multirow{4}{*}{RaFL (Ours)}              & \multirow{2}{*}{ResNet} & \multirow{4}{*}{CIFAR-100}            & 47 MFLOPs                                                                                              & \multirow{4}{*}{100} & 0.66 GB              & \textbf{340 GB}      & \textbf{(62.01$\pm$1.84)\%}  & \textbf{+22.01\%}  \\
                                          &                         &                                       & 74 MFLOPs                                                                                              &                      & 0.80GB               & \textbf{400GB}       & \textbf{(65.28$\pm$1.22)\%}  & \textbf{+23.07\%}  \\
                                          & MobileNet-V2            &                                       & 12 MFLOPs                                                                                              &                      & 180MB                & \textbf{88GB}        & \textbf{(58.66$\pm$1.40)\%}  & \textbf{+30.86\%}  \\
                                          & MobileNet-V3            &                                       & 6.5 MFLOPs                                                                                             &                      & 140MB                & \textbf{68GB}        & \textbf{(50.49$\pm$2.49)\%}  & \textbf{+24.6\%}   \\
\hline
\end{tabular}
}
\vspace{-.6cm}
\end{table*}


\subsection{Resource-aware System Heterogeneity}
\label{sec:resource_hete}
\begin{table}[ht]
\centering
\caption{Resource utilization efficiency}
\label{tab:resource_aware}
\resizebox{.55\columnwidth}{!}{%
\begin{tabular}{cccccc}
\hline
\textbf{Method} &
  \textbf{Model} &
  \textbf{\begin{tabular}[c]{@{}c@{}}Total \\ Resource\end{tabular}} &
  \textbf{\begin{tabular}[c]{@{}c@{}}Utilized \\ Resources\end{tabular}} &
  \textbf{\begin{tabular}[c]{@{}c@{}}Resource\\ Utilization\end{tabular}} &
  \textbf{Accuracy} \\ \hline

\multirow{4}{*}{\begin{tabular}[c]{@{}c@{}}Uniformed\\ Baselines\end{tabular}} &
  \multirow{2}{*}{ResNet} &
  5GFLOPs &
  2.75GFLOPs &
  55\% &
  41\% \\
                      &                         & 8GFLOPs   & 5.02GFLOPs & 63\%          & 43\%          \\
                      & MobileNet-V2            & 1.2GFLOPs & 1.07GFLOPs & 89\%          & 29\%          \\
                      & MobileNet-V3            & 0.7GFLOPs & 0.49MFLOPs & 70\%          & 27\%          \\ \hline

\multirow{4}{*}{RaFL} & \multirow{2}{*}{ResNet} & 5GFLOPs   & 4.63GFLOPs & \textbf{93\%} & \textbf{65\%} \\
                      &                         & 8GFLOPs   & 7.26GFLOPs & \textbf{91\%} & \textbf{67\%} \\
                      & MobileNet-V2            & 1.2GFLOPs & 1.15GFLOPs & \textbf{99\%} & \textbf{60\%} \\
                      & MobileNet-V3            & 0.7GFLOPs & 0.64GFLOP  & \textbf{91\%} & \textbf{57\%} \\ \hline
\end{tabular}%
}
\vspace{-.8cm}
\end{table}
We investigated \proj's ability to cope with resource heterogeneity by evaluating its resource utilization efficiency and learning efficiency under system heterogeneity.

\noindent\textbf{Resource Utilization Comparison.} Compared with uniform model deployment FL methods (such as FedAvg~\cite{mcmahan2017fedavg}, FedNova~\cite{wang2020fednova}, FedProx~\cite{li2020fedprox}, SCAFFOLD~\cite{karimireddy2020scaffold}), Table~\ref{tab:resource_aware} shows that \proj demonstrates significant resource utilization efficiency and high accuracy overhead. At least $90\%$ of the resources were utilized by \proj across different resource constraints, with the highest utilization being $99\%$. In contrast, the single model deployment in the baselines would have their resource utilization limited by the clients with the lowest resource capacity.

\noindent\textbf{Model Performance under Different Resource Budget.} Figure~\ref{fig:resource_heto} illustrates the impact of different resource budgets on model accuracy optimized by \proj. \proj exhibits stable performance under various resource heterogeneity scenarios. Even with significant differences in preset resource overhead, \proj maximizes learning efficiency and resource efficiency, producing stable final model performance. This adaptability and effectiveness highlight \proj's ability to address diverse resource constraints while maintaining consistent performance across various systems.

\begin{figure*}[t]
 \begin{center}

\centerline{
\includegraphics[width=.25\columnwidth]{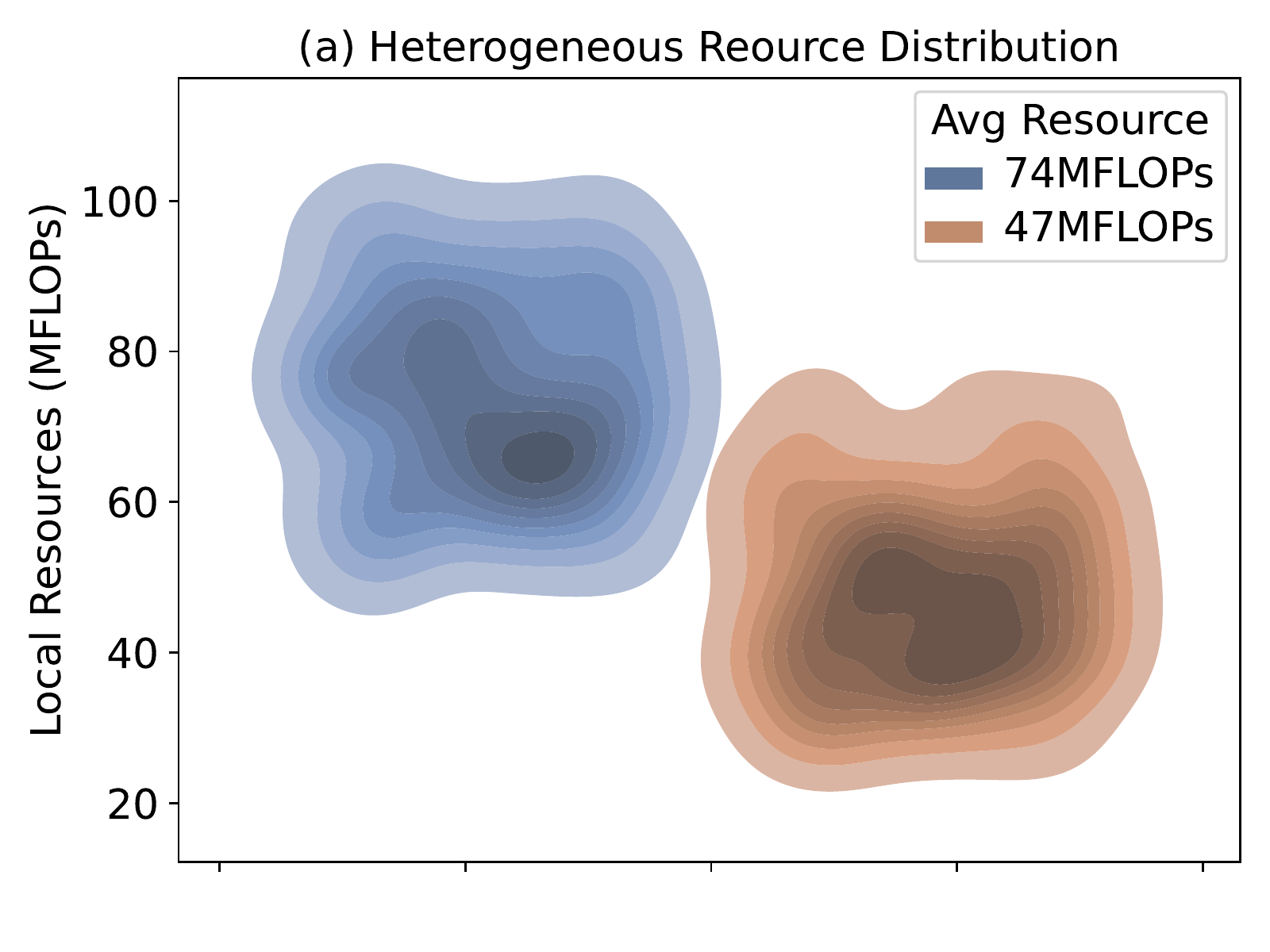}
\includegraphics[width=0.25\columnwidth]{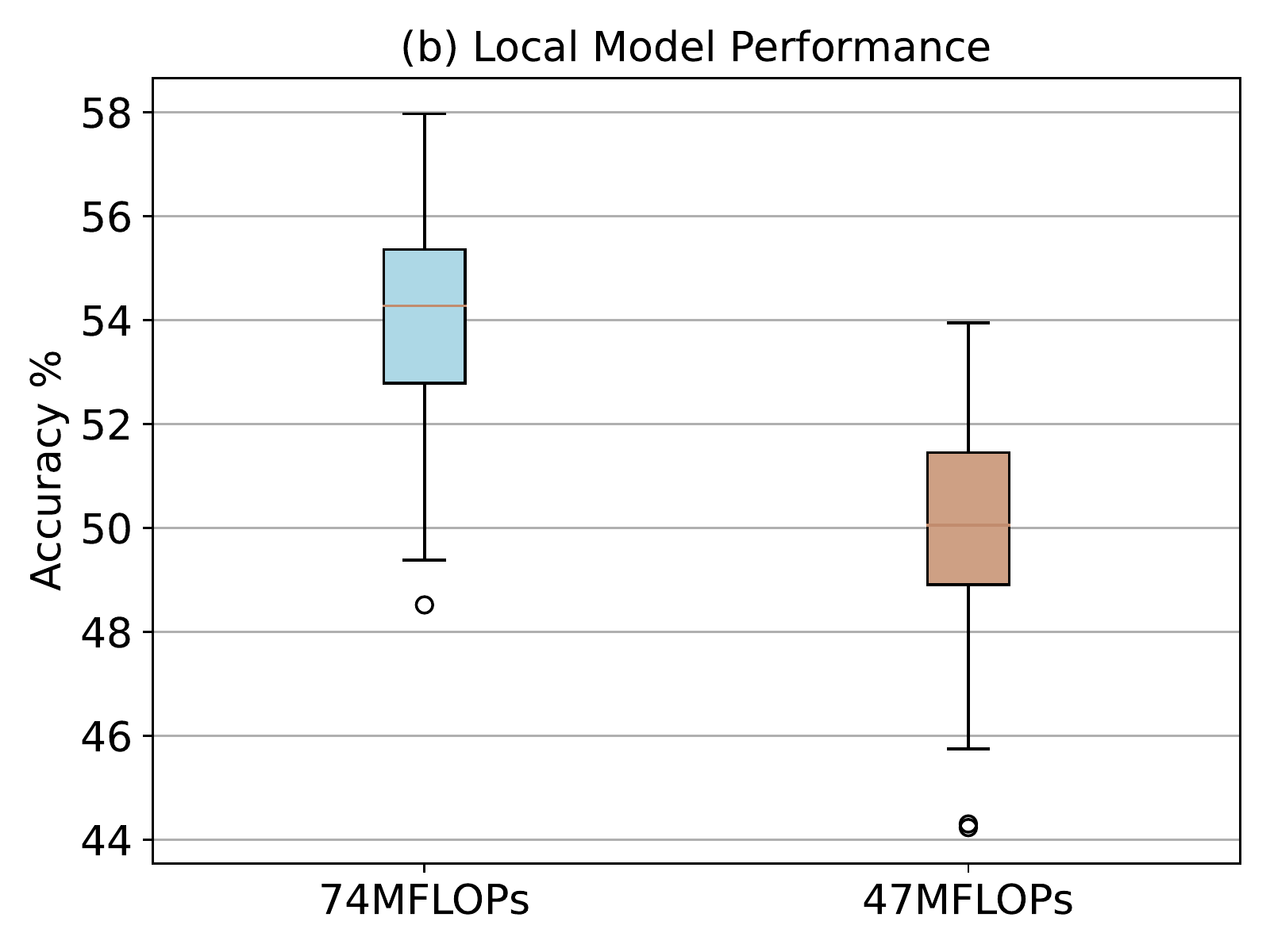}
}\vspace{-.4 cm}
\caption{Local model performance in resource heterogeneous FL.}
\label{fig:resource_aware}
\end{center}
 \end{figure*}

\noindent\textbf{Edge Performance under System Heterogeneity.} We set up two systems with different average resource constraints (74MFLOPs and 47MFLOPs). Figure~\ref{fig:resource_aware} (a) shows the heterogeneous resource distributions within these systems. Despite the significant variance in local client resources, Figure~\ref{fig:resource_aware} (b) demonstrates that the models running at the edge exhibit similar performance. This can be attributed to the optimal architecture search performed by the neural architecture search under the provided resource constraints, emphasizing \proj's effectiveness in addressing system heterogeneity and ensuring consistent performance across diverse edge devices.

\subsection{Scaled and Sporadic FL}
\begin{figure}[t]
 \begin{center}
\vspace{-1cm}
\centerline{
\includegraphics[width=.5\columnwidth]{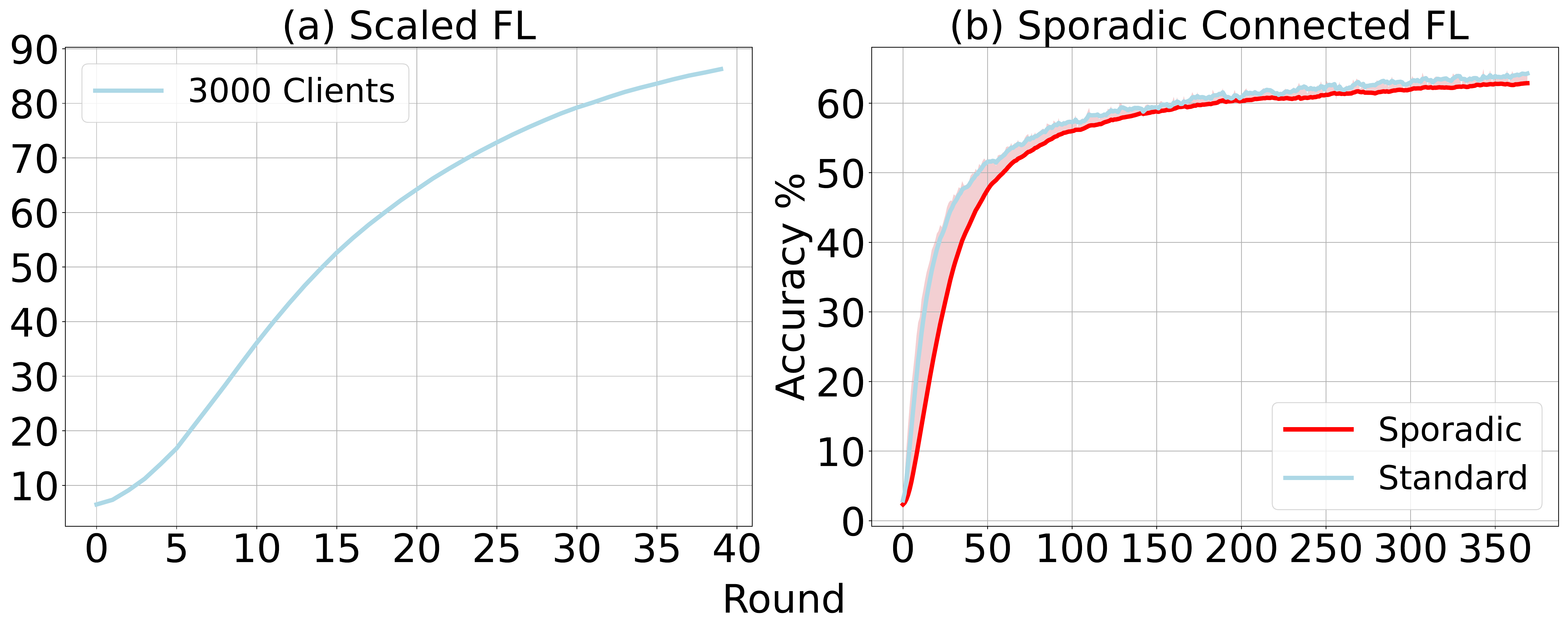}
}\vspace{-.4 cm}
\caption{Scaled and Sporadic Connected FL.}
\vspace{-1.25 cm}
\label{fig:scaled}
\end{center}
 \end{figure}
In practice, a typical FL system interacts with a large number of clients. The system is dynamic and asynchronous, meaning new clients may join and leave the network at any stage of FL training, and the number of participating clients may be large.
Figure~\ref{fig:scaled} shows \proj's performance under scaled and sporadic FL settings. Large-scale federated learning is one of the biggest challenges in FL, as shown in Figure~\ref{fig:scaled} (a), we set up 3000 clients with distinct heterogeneous local data on FEMNIST, \proj converges quickly and remains stable.
Figure~\ref{fig:scaled} (b) shows the sporadic FL in CIFAR-100. We implemented a client stream to simulate a sporadic FL setting. For each round of communication, we uniformly replace $10\%$ of clients with new ones, to mimic 10\% of clients losing connections and 10\% of clients joining the current stage of training. \proj achieves impressive results under sporadic FL and produces competitive model performance compared to the standard FL setting. This is despite the total clients flowing through the sporadic FL setting equalling 3500, which is $35\times$ more than the standard setting. The results highlight \proj's robustness and adaptability in dynamic and large-scale federated learning scenarios.

\subsection{Comparison with FL approaches that use Neural Architecture Search}
\noindent\textbf{Learning Efficiency Comparison.} We compared \proj to FedNAS~\cite{he2020fednas}, which deploys super-networks directly to edge devices and searches sub-networks locally for edge applications. Figure~\ref{fig:overfitting}(b) shows that FedNAS diverged due to overfitting in our experiment settings. We also analyzed DecNAS~\cite{xu2020decnas}.
It is important to note that \proj has a different objective compared to FedNAS~\cite{he2020fednas} and DecNAS~\cite{xu2020decnas}. While FedNAS and DecNAS aim to optimize the NAS super-network via federated learning on user private data, \proj focuses on efficiently deploying resource-aware specialized networks to heterogeneous clients to maximize resource utilization and achieve robust performance in dynamic and large-scale federated learning scenarios.

\noindent\textbf{Distinguishing RaFL from Other NAS-based Federated Learning Approaches.} Other approaches may not yield efficient neural architectures. For instance, FedNAS performs neural architecture search at the beginning of the training, which may not result in high-performance architectures.
Existing methods might not effectively tackle resource heterogeneity in edge devices. For example, FedNAS conducts neural architecture search directly on edge devices, which can be computationally demanding and unsuitable for resource-limited edge devices.
Optimizing super-networks in FL environments can easily lead to divergence. These approaches might work when the number of clients is limited. However, when the number of clients in FL becomes large, it becomes almost impossible to converge the over-parameterized super-networks. In contrast, \proj aims to efficiently deploy resource-aware specialized networks to heterogeneous clients to maximize resource utilization and achieve robust performance in dynamic and large-scale federated learning scenarios.

RaFL ingeniously avoids the above limitations by dynamically deploying specialized high-performance networks to edge clients based on their local resource overhead, enabling efficient utilization of local resources and better downstream tasks.


\subsection{Comparison with KD-based FL}
\label{sec:kdbaselines}
As shown in Table~\ref{tab:kd_baselines}, utilizing knowledge distillation to aggregate local models may not enhance the learning efficiency of the FL system. The performance of knowledge distillation heavily relies on the similarity between public data and local data, making it unsuitable for all general FL environments.
We compared \proj with the knowledge distillation baseline FedDF~\cite{lin2020feddf} under various public data scenarios. \proj is more robust; it initializes the global knowledge network by weighted averaging local knowledge networks using Equation~\ref{eq:aggre} and then optionally performs ensemble knowledge distillation. In cases where the public data significantly deviates from the overall local data distribution in FL, \proj experiences fewer negative effects.

\subsection{Extra Burdens of Knowledge Networks}
\label{sec:extra_cost}

To address concerns regarding the potential extra computational overhead introduced by incorporating knowledge networks and performing local deep mutual learning in the \proj framework, we designed and conducted experiments using two distinct FL environments:
Environment A: 12 MFLOPs MobileNet-V3 and an 8 MFLOPs MobileNet-V3 knowledge network deployed at each local client.
Environment B: 20 MFLOPs MobileNet-V3 deployed at each local client, directly updating the model with the SGD algorithm without knowledge distillation.
Both FL environments were trained for 500 rounds separately using the same nodes and the same type of GPU (NVIDIA V100). Results show that Environment B required 49 hours and 19 minutes to complete training, while Environment A only took 48 hours and 25 minutes. This indicates that incorporating knowledge networks and performing deep mutual learning did not add significant computational overhead to the edge.
\begin{table}[]
\vspace{-1cm}
\centering
\caption{Comparison with FedDF under different public data.}
\label{tab:kd_baselines}
\resizebox{.5\columnwidth}{!}{%
\begin{tabular}{llllll}
\hline
\textbf{Public Data} & \textbf{Method} & \textbf{\begin{tabular}[c]{@{}l@{}}Local \\ Dataset\end{tabular}} & \textbf{Clients} & \textbf{Participation} & \textbf{Accuracy} \\ \hline
CINIC-10     & FedDF    & CIFAR-100 & 100 & 10\% & diverge \\
             & RaFL-(b) &           & 100 & 10\% & 50.51   \\ \hline
CIFAR-10     & FedDF    & CIFAR-10  & 30  & 40\% & 73.35   \\
             & RaFL-(b) &           & 30  & 40\% & 72.49   \\ \hline
TinyImageNet & FedDF    & CIFAR-100 & 100 & 10\% & diverge \\
             & RaFL-(b) &           & 100 & 10\% & 54.95   \\ \hline
\end{tabular}%
}
\vspace{-1.2cm}
\end{table}

\section{Conclusion}
\label{sec:con}
In this paper, we present \proj, a resource-aware FL approach that seamlessly integrates AutoML solutions, such as NAS, into FL environments. By providing clients with specialized networks tailored to their individual resource constraints, \proj offers a robust solution for achieving efficient learning and improved resource utilization.
Experiments conducted on various heterogeneous FL environments demonstrate \proj's practicality and efficacy in enabling heterogeneous FL with faster convergence and enhanced communication efficiency. The promising results obtained by \proj highlight the potential of integrating AutoML solutions into FL environments to better accommodate the diverse resource constraints of edge devices, paving the way for wider applications in real-world situations. Moreover, \proj exhibits effectiveness in addressing FL scenarios, showcasing its potential for easy extension to decentralized and distributed training frameworks.

\section*{Acknowledgment}
This research was supported by the National Science Foundation under Grant number 2243775. We also appreciate the provision of computational resources by Intel Labs for this project. Furthermore, we thank the ResearchIT team \footnote{https://researchit.las.iastate.edu} at Iowa State University for their continuous and helpful assistance.

\bibliography{example_paper}
\bibliographystyle{abbrv}

\newpage
\appendix
\onecolumn



\end{document}